\documentclass[10pt,twocolumn,letterpaper]{article}

\usepackage{cvpr}              % To produce the CAMERA-READY version
\usepackage[dvipsnames]{xcolor}

\definecolor{cvprblue}{rgb}{0.21,0.49,0.74}
\usepackage[pagebackref,breaklinks,colorlinks,citecolor=cvprblue]{hyperref}
\usepackage{amsmath}

\def\paperID{34} % *** Enter the Paper ID here
\def\confName{CVPR}
\def\confYear{2024}

\title{Sketch-guided Image Inpainting with Partial Discrete Diffusion Process}

\author{Nakul Sharma$^{1}$,~~~~Aditay Tripathi$^{2}$,~~~~~Anirban Chakraborty$^{2}$,~~~~Anand Mishra$^{1}$\\
$^1$IIT Jodhpur~~~~~~~~~~$^2$IISc, Bengaluru\\
{\tt\small sharma.86@iitj.ac.in, aditayt@iisc.ac.in, anirban@iisc.ac.in, mishra@iitj.ac.in}
}

\begin{document}
\twocolumn[{%
\renewcommand\twocolumn[1][]{#1}%
\maketitle
%\begin{center}
%    \centering
%    \captionsetup{type=figure}
%    \includegraphics[width=\textwidth]{Figures/teaser_figure_1row.pdf}
%    \captionof{figure}{\label{fig:problem_statement} \textbf{Sketch-guided Image Inpainting:}
%     Columns 1-4 show the original image, masked image, sketch query, and inpainting using our approach that leverages sketch.
%    The proposed sketch-guided image inpainting is capable of utilizing the shape and pose of an object provided by a crude sketch to fill in the missing region with high visual fidelity and sketch faithfulness. In contrast, the results obtained using unconditioned inpainting methods fail to achieve comparable quality.}

%\end{center}%
}]

\begin{abstract}
   In this work, we study the task of sketch-guided image inpainting. Unlike the well-explored natural language-guided image inpainting, which excels in capturing semantic details, the relatively less-studied sketch-guided inpainting offers greater user control in specifying the object's shape and pose to be inpainted. As one of the early solutions to this task, we introduce a novel \textit{partial discrete diffusion process} (PDDP). The forward pass of the PDDP corrupts the masked regions of the image and the backward pass reconstructs these masked regions conditioned on hand-drawn sketches using our proposed sketch-guided bi-directional transformer. The proposed novel transformer module accepts two inputs -- the image containing the masked region to be inpainted and the query sketch to model the reverse diffusion process. This strategy effectively addresses the domain gap between sketches and natural images, thereby, enhancing the quality of inpainting results. In the absence of a large-scale dataset specific to this task, we synthesize a dataset from the MS-COCO to train and extensively evaluate our proposed framework against various competent approaches in the literature. The qualitative and quantitative results and user studies establish that the proposed method inpaints realistic objects that fit the context in terms of the visual appearance of the provided sketch. To aid further research, we have made our code publicly available here: \url{https://github.com/vl2g/Sketch-Inpainting}.
\end{abstract}
    
\section{Introduction}
\label{sec:intro}
\begin{figure}[!t]
    \centering
    \includegraphics[width=0.5\textwidth]{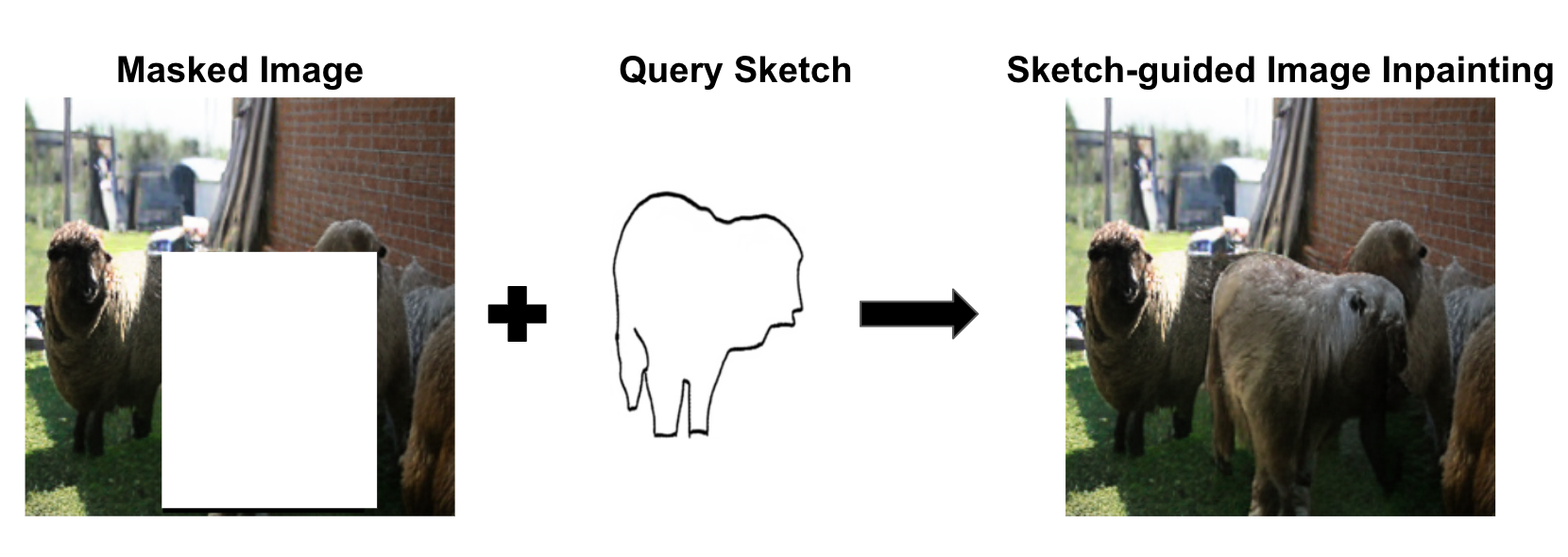}
    \caption{\textbf{Sketch-guided Image Inpainting} has been an under-explored task in the literature and is often restricted to partial sketch-based image manipulation~\cite{jo2019sc, deepfillv2, deepfillv2}. We fill this gap in the literature by proposing a novel partial discrete diffusion process for sketch-guided object-level inpainting. Our proposed approach significantly outperforms other plausible approaches on Sketch-guided Image Inpainting.}
    \label{fig:goal}
\end{figure}

Image inpainting is a well-established task in computer vision with diverse applications, including natural photo editing~\cite{deepfillv2, suvorov2022largeholeinpainting, 2022RePaintIU} and filling missing data in medical images~\cite{MedGAN, multi-task_med_inp, InpMedModalities}.
Significant progress has been made in image painting in recent years, partly thanks to large neural models~\cite{2022GLaMaJS, Chang2022MaskGITMG, MaskAwareTransformer}. Despite remarkable progress, most current image inpainting methods rely solely on available image regions as context for inpainting. Consequently, these ``unconditioned image inpainting methods''  lack precise control over semantic object categories or visual attributes such as object shape and pose within the targeted inpainting region. Imagine a scenario where a user intends to inpaint a specific object with precise characteristics like it's size, pose, and shape in the masked image region. While one intuitive approach is to use natural language descriptions for text-guided image inpainting akin to~\cite{zhang2020text-guided}, guiding image inpainting using sketch emerges as a promising alternative, especially for users with stronger artistic abilities than linguistic skills.
Further, as noted in fine-grained Sketch-based Image Retrieval literature~\cite{BhuniafinegrainSBIR,li2014fine}, an important characteristic of sketches lies in their ability to capture object appearance and structure intrinsically. This motivates us to propose and study the \emph{sketch-guided image inpainting} as an independent problem. While certain prior studies~\cite{deepfillv2, Liu2021DeFLOCNet, scfegan} incorporate partial sketch information for \emph{image manipulation} akin to inpainting, our task distinctively focuses on object-level inpainting utilizing complete object sketches instead of partial sketch strokes. Our goal and a selected result of our approach are illustrated in Figure~\ref{fig:goal}. 

Sketch-guided image inpainting requires precise utilization of object-level shape and pose information conveyed by the query sketch while dealing with the large domain gap between hand-drawn sketches and images. We approach this problem by proposing a novel method based on the discrete diffusion process~\cite{2021DiscreteDiffusion} to inpaint the missing regions conditioned on the hand-drawn query sketch. Our approach involves two main stages: In the first stage, we learn a visual codebook to describe a discrete latent space of images. This codebook enables us to represent images compactly. In the second stage, we model the sketch-guided image inpainting problem in this discrete latent space. Here, we introduce a novel \emph{Partial Discrete Diffusion process} or PDDP tailored for sketch-guided inpainting, which allows for controlled corruption of the image region and subsequent reconstruction guided by the query sketch. Specifically, we propose a \emph{sketch-guided bi-directional transformer} model to reverse the diffusion process, thereby effectively inpainting the missing regions based on the provided sketch. During inference, the inpainting process reduces to reversing the partial discrete diffusion process for the masked area of the image, guided by the user-provided sketch. This enables our method to generate high-quality inpainted images that faithfully capture the intended object shapes and poses specified by the input sketches. 

Given the absence of a suitable dataset that can be used to study sketch-guided image inpainting, we curated a dataset specifically for this task by leveraging the rich annotations available in the MS-COCO dataset~\cite{cocodataset}. Specifically, we segment out objects from the images and sketchify them using an off-the-shelf model~\cite{photosketching}. We perform extensive experiments on our curated dataset and compare existing image inpainting approaches adapted for our task with ours. %Our proposed approach achieves superior visual fidelity, semantic consistency, and sketch faithfulness, highlighting its efficacy in addressing the unique challenges posed by sketch-guided image inpainting. 

In summary, our contributions are as follows: (i) We study the task of sketch-guided image inpainting, which involves completing the missing region in an image while considering the shape and pose details of the object provided in the accompanying hand-drawn sketch. This work can be seen as a first attempt at studying sketch guidance in image painting at the object level. (ii) To tackle this challenging problem, we first learn the latent representation of the natural images and model the forward diffusion process only on the masked image regions. We propose the {Partial Discrete Diffusion Model} to learn the sketch conditional reverse diffusion process to complete the image by incorporating the visual information from the provided hand-drawn sketch. (iii) We compare the performance of our model with suitable baselines and establish a new state-of-the-art for our proposed sketch-guided image inpainting task. We have made our code publicly available here: \url{https://github.com/vl2g/Sketch-Inpainting}.
%firstly, due to the unavailability of suitable datasets tailored for sketch-guided inpainting, we curated a dataset specifically for this task by leveraging the rich annotations available in the MS-COCO dataset~\cite{cocodataset}. Specifically, we segment out objects from the images and create their synthetic sketch version using an off-the-shelf model ~\cite{photosketching}. Next,
\section{Related Work}
In recent years, we have seen rapid progress in the deep learning applications for the sketch domain; these include sketch-based image retrieval~\cite{BhuniafinegrainSBIR,li2014fine}, object localization~\cite{tripathi2020sketch, Tripathi_2024_WACV_Sketch},  sketch generation~\cite{wang2023sketchknitter}, scene-level sketch-based image retrieval~\cite{scene-level}, etc. In this work, we shall focus on inpainting and sketch-to-image generation literature.\\ 
\noindent\textbf{Image Inpainting:}
Traditional approaches of image inpainting rely on propagating low-level features from surrounding image content to reconstruct the missing regions~\cite{00imageinpainting, 2004inpaintingcriminisi}. More recently, deep learning-based methods have significantly progressed by leveraging semantic image representations learned by convolutional neural networks.
Context encoders~\cite{pathak2016context} introduced an encoder-decoder architecture to generate the contents of an irregularly shaped hole based on the surrounding image context. Subsequent works have expanded on this approach with attention mechanisms~\cite{yu2018generative}, adversarial training~\cite{deepfillv2}, and improved network architectures~\cite{nazeri2020edgeconnect}.  Another paradigm leverages diffusion models, which can synthesize high-quality images by learned reverse diffusion processes~\cite{ho2020ddpm, 2021PaletteID}. RePaint~\cite{2022RePaintIU} exploits pre-trained unconditional DDPMs~\cite{ho2020ddpm} to improve the inpainting process by diffusion models.

Guided image inpainting began with Zhang et al.'s work~\cite{zhang2020text-guided} via TDANet, employing a dual attention mechanism to utilize textual cues for inpainting by comparing text with the corrupted and original images. Diffusion-based text-to-image generation models like~\cite{ramesh2021zero, ding2021cogview, nichol2021glide, ramesh2022dalle2, 2021LatentDiffusion, 2021VectorQD} can be used directly for text-guided image inpainting; however, these methods can produce sub-optimal results as they are trained on image-level captions instead of object instance-specific descriptions. Recently, ~\cite{ni2023nuwa-lip} proposes a text-guided image inpainting framework leveraging a defect-free VQ-GAN version for improved inpainting results. Zeng et al.~\cite{zeng2022shape} proposed shape-guided object inpainting in images and subsequently, ~\cite{xie2023smartbrush, Park_2024shape-inside-out-attn} proposed frameworks to guide the inpainting process using the shape of the mask region along with the text. Our problem setup involves providing a user sketch of an object which should be inpainted. This gives the user better control over specifying the object's shape, pose, and size.
\newline
\noindent\textbf{Sketch-to-Image Generation:}
Sketch2Photo~\cite{sketch2photo} and PhotoSketching~\cite{photosketching} synthesized whole images by compositing the retrieved foreground and background images using a given sketch. Gao et al.~\cite{gao2020sketchycoco} introduced a two-stage method using EdgeGAN to generate realistic images from scene-level sketches. Initial works on object-level image generation from sketches include~\cite{chen2018sketchygan, lu2018contextualgan}. More recent works AODA~\cite{xiang22advopen} and~\cite{koley23picthtsktch} propose methods for open-set object-level image generation from sketches.
With the rise of text-to-image diffusion models, interest has been in controlling the generation using sketches. Voynov et al.~\cite{voynov2023sketchguidedt2i} trains a latent-guided predictor module that maps latent features of noisy images to spatial maps for providing sketch-guidance to text-to-image diffusion models. ControlNet~\cite{controlnet23} utilizes pre-trained StableDiffusion~\cite{2021LatentDiffusion} by learning parallel architectures for different modalities (edge, scribble, depth maps, pose, etc.) and uses it along with modality-specific guidance to control the image generation process.ControlNet can guide the StableDiffusion model using scribbles for scene-level image synthesis. Since diffusion models are capable of doing inpainting inherently~\cite{ho2020ddpm, nichol2021glide, 2021PaletteID}, we can utilize ControlNet for our task as a plausible approach by providing sketch input of only the region that needs to be inpainted. We experimentally compare against such a baseline in Section~\ref{sec:expts}.\\
\noindent\textbf{Sketch-based Image Manipulation:}
Previous research on sketch-based image manipulation is primarily based on a conditional image inpainting framework. For instance, DeepFill-v2 \cite{deepfillv2} enables the manipulation of both general and facial images via partial sketches. Meanwhile, FaceShop~\cite{portenier2018faceshop} permits localized shape and color adjustments in facial images through sketch-based manipulation accomplished via conditional image completion. SC-FEGAN \cite{jo2019sc} delves into facial manipulation via sketches and color strokes by integrating free-form masks and style loss into the image completion model. Yang et al.~\cite{iizuka2017globally} adapt a face manipulation model trained on sketches generated by edge detection to human-drawn sketches, and propose a refinement strategy that dilates and refines user-drawn sketches to resemble edge detection results. These methods combine an input image and these low-level controls for CNN inputs. However, the corresponding feature representations are not sufficient to convey user intentions. DeFLOCNet~\cite{Liu2021DeFLOCNet} deals with this problem by proposing a new architecture capable of preserving these control features in the deep feature representations. SketchEdit~\cite{zeng2022sketchedit} introduces a mask-free image manipulation framework using partial strokes. While our work draws inspiration from these methods, we focus on object-level inpainting. %Specifically, our problem focuses on inserting an object in the masked region described crudely by the user-provided sketch.
\begin{figure*}[h]
% stage-2: https://docs.google.com/drawings/d/1z5rkTlM_y_wxHSyzhbDgzCmUzsnRzVIWKqctSpGq9bA/edit?usp=sharing
% stage-1: https://docs.google.com/drawings/d/1fE0IyzKj47hsFFJYxDKZAovrXW6eD1UYhd44u_Xny9k/edit?usp=sharing
\centering
\includegraphics[width=\textwidth]{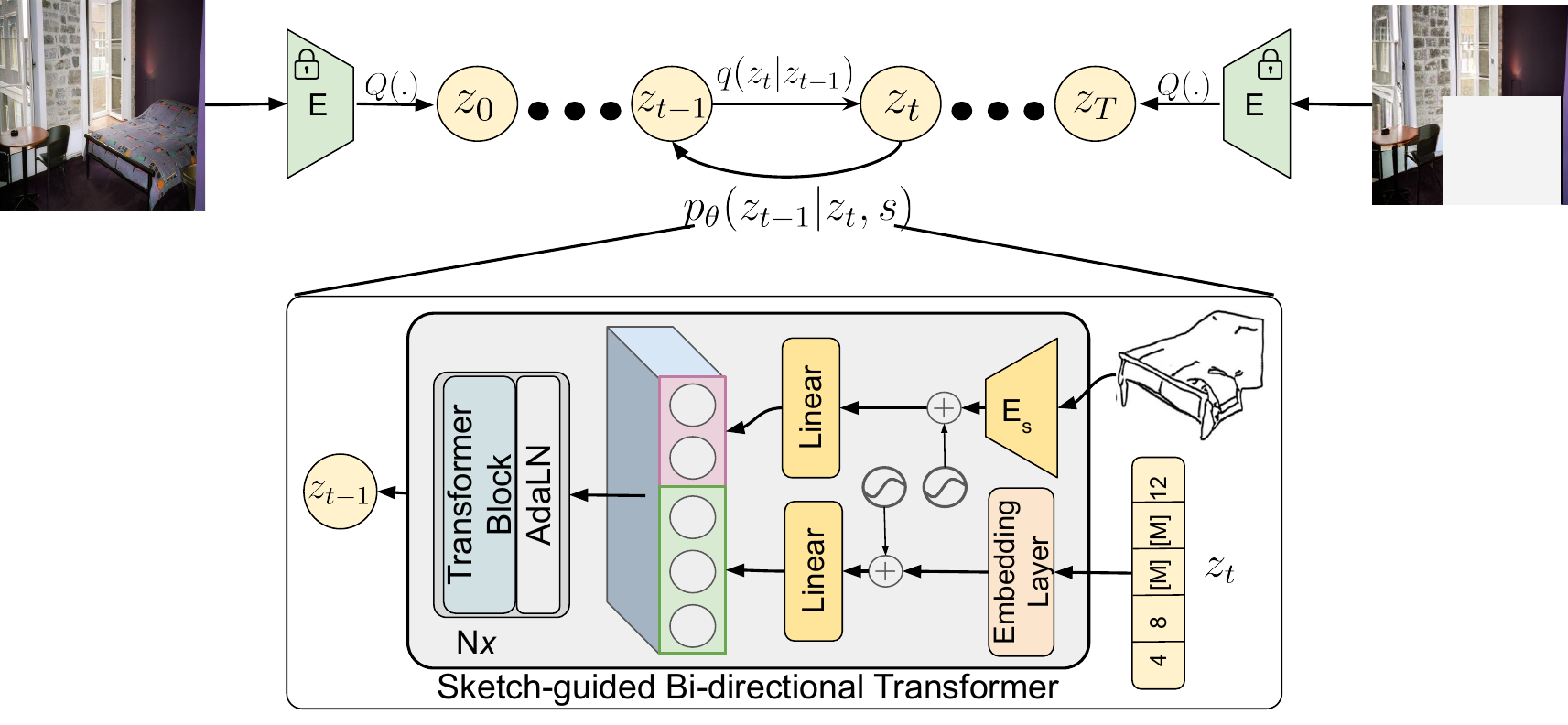}
\caption{\label{fig:model_figure} Our method involves obtaining a discrete latent space representation of the original image and its masked counterpart using a pretrained VQ-VAE. The image is first converted to noise by iteratively adding noise to the masked region in the forward process of the proposed Partial Discrete Diffusion Process (Section~\ref{subsec:discrete_diff}). In the reverse process, sketch-guided inpainting is performed iteratively using a sketch-guided bi-directional transformer model that takes the masked image tokens and the query sketch. It predicts the tokens of the missing regions (Section~\ref{subsubsec:reverse_process}). By iteratively refining the inpainted image using the sketch and the available information from the original image, the proposed method can generate high-quality inpainted images with correct visual and pose details. (Best viewed in color).}
\end{figure*}

\section{Methodology}
Given an image $I$ with missing regions defined by a mask $M$ and a hand-drawn sketch $S$ containing high-level details of an object's appearance, our objective is to generate a completed image $I'$ that contains the generated object that precisely follows the visual, e.g., shape, pose information provided in the hand-drawn sketch. To achieve this, a two-stage methodology is proposed in this work. In the first stage, images are represented as a sequence of codebook indices following~\cite{Esser2020TamingTF}. In the second stage,  the codebook representation of images is utilized to inpaint using a partial discrete diffusion approach conditioned on the hand-drawn sketches. In the next section, we formally introduce the problem statement of sketch-guided image inpainting (Section~\ref{sec:problem_statement}), briefly explain the discrete diffusion model (Section~\ref{subsec:discrete_diff}), and finally describe our approach (Refer to Section~\ref{sec:model}) to address the problem.

\subsection{Problem Setup}
\label{sec:problem_statement}
Let $I \in \mathbb{R}^{H \times W \times 3}$ be a three-color channel input image, where $H$ and $W$ are the height and width of the image, respectively. Further, let $M \in \mathbb{R}^{H \times W \times 1}$ be a binary mask that indicates the missing regions of the input image. Each element $M_{ij} \in \{0,1\}$ on the binary mask $M$ represents whether the pixel at the location $(i,j)$ on the image $I$ is missing. Let $S \in \mathbb{R}^{H_s \times W_s \times 1}$ be a free-hand sketch that provides high-level visual guidance for the appearance and pose of an \emph{object} in the missing regions. In the proposed sketch-guided image inpainting task, we aim to learn a generative model $G$ that takes an image $I$ with the missing region represented by the mask $M$ and a hand-drawn sketch $S$ as input and outputs a completed image $I' \in \mathbb{R}^{H \times W \times 3}$ such that the object inpainted in the missing region is visually consistent with other regions of the image and is visually coherent with the provided hand-drawn sketch. 
% Specifically, we want to minimize the difference between the missing regions in $I'$ and the ground truth image $I_{GT}$.
\subsection{Preliminary: Discrete Diffusion}
\label{subsec:discrete_diff}
% Diffusion models \cite{sohl15diffusion} are a type of generative model that involve latent variables and are distinguished by a forward and reverse Markov process. The forward process corrupts the data sample iteratively for a fixed number of timesteps by adding Gaussian noise sampled from a timestep-dependent distribution. During the reverse process, we train a neural network that learns to denoise each step to the data distribution.
% The discrete diffusion models were first introduced by Sohl-Dickstein et al. \cite{sohl15diffusion}. 
D3PM~\cite{2021DiscreteDiffusion} introduced a general diffusion framework in discrete space for categorical variables. We will first describe the forward diffusion process for a discrete diffusion model with total time-steps $T \in \mathbb{N}$. For a discrete random variable at time $t\in [1, T]$, $z_t \in \{1, 2, 3, \ldots , C-1, C\}$, the transition matrix $\mathbf{Q_t} \in [0, 1]^{C \times C}$ defines transition probabilities associated with each state that $z_t$ can take. More formally, it defines the probabilities that $z_{t-1}$ transits to $z_t$, $[Q_t]_{mn} = q(z_t = m \lvert z_{t-1} = n)$ in a single time step. It is mathematically described as $q(z_t \lvert z_{t-1}) = {v}(z_t)^\top\mathbf{Q_t}{v}(z_{t-1}),$
% \begin{center}
%     \begin{equation}
%         q(z_t \lvert z_{t-1}) = \vb*{v}(z_t)^\top\mathbf{Q_t}\vb*{v}(z_{t-1}),
%     \end{equation}
% \end{center} 
where ${v}(.)$ denotes a function that encodes a nominal value to one-hot vector over $C$ categories, i.e., ${v}(z) \in \{0, 1\}^{C}$. Assuming the Markov property, the $t$ step transition probabilities can be obtained as $q(z_t \lvert z_{0}) = {v}(z_t)^{\top}\mathbf{\bar{Q}_t}{v}(z_{0})$,
% \begin{center}
%     \begin{equation}
%         q(z_t \lvert z_{0}) = \vb*{v}(z_t)^{\top}\mathbf{\bar{Q}_t}\vb*{v}(z_{0}),
%     \end{equation}
% \end{center} 
where $\mathbf{{\bar{Q}}_t} = \prod_{i=t}^{1} \mathbf{Q_{i}}$.  This analysis can be extended to the N-dimensional random variables $z_t \in \{1, 2, 3, \ldots, C\}^N$, and the transition matrix is used for each variable in the random vector $z_t$ independently. From here onwards, we consider $z_t$ as an N-dimensional random variable representing N discrete tokens. D3PM, inspired by the masked language modeling task in NLP, proposes \emph{absorbing state} formulation of this transition matrix and introduces a {\tt [MASK]} token and argues that this special token helps identify corrupted and non-corrupted regions. VQ-Diffusion~\cite{2021VectorQD} advances this formulation of matrix $\mathbf{Q_t}$ with their mask-and-diffuse strategy for image generation by introducing three probabilities $\gamma_t$ of replacing the current token with the {\tt [MASK]} token, $\beta_t$ of replacing the current token with another token, and $\alpha_t$ describing the probability of token to retain its state. The transition matrix $\mathbf{Q_t} \in [0, 1]^{(C+1) \times (C+1)}$ is then given by:
\begin{center}
    \begin{equation}
    \label{eqn:Q_eqn}
    \scriptsize
\mathbf{Q}_{t} = \begin{pmatrix}
\alpha_{t}+\beta_{t} & \beta_{t} & \ldots & \beta_{t} & 0  \\
\beta_{t} & \alpha_{t}+\beta_{t} & \ldots & \beta_{t} & 0  \\
\vdots & \vdots & \ddots & \vdots & \vdots  \\
\beta_{t} & \beta_{t} & \ldots & \alpha_{t}+\beta_{t} &  0 \\
\gamma_{t} & \gamma_{t} & \ldots & \gamma_{t} & 1 \\
\end{pmatrix}.
\end{equation}
\end{center}

The reverse process is parameterized by a neural network that models the $z_{t-1}$ distribution given $z_t$. Specifically, $z_{t-1}$ is sampled from $p_\theta(z_{t-1}|z_t) \in [0, 1]^{(C+1) \times (C+1)}$. During inference, image synthesis tasks initialize all tokens of $z_T$ as {\tt [MASK]}, and then iteratively sample denoised latents from $p_\theta(z_{t-1} \lvert z_t)$ to obtain $z_0$~\cite{2021DiscreteDiffusion, 2021VectorQD}.

\subsection{Sketch-guided Image Inpainting using Partial Discrete Diffusion Process}
\label{sec:model}
This section describes the two-stage model we have designed for the sketch-guided image inpainting task. In the first stage, we train an encoder $E$, a codebook $Z$, and a decoder $D$ to learn a perceptually compressed discrete latent space of images using the method described by Esser et al.~\cite{Esser2020TamingTF}. Any image can then be represented as a sequence of indices of latent vectors from the codebook as $z_0 \in \{1, 2, \ldots, C\}^K$, where $K$ denotes the number of visual tokens representing the image in the discrete space. Our novelty lies in the second stage where we first project the ground truth image and the masked image in the discrete latent space using the learned encoder and the codebook and then use these discrete representations in the discrete diffusion process. For each ground truth image $I_{GT} \in \mathbb{R}^{H \times W \times 3}$, we randomly mask an object in the image using mask $M \in \{0,1\}^{H \times W \times 1}$ where $M(i,j)=1$ means that the image regions $I_{GT}(i,j)$ is masked. Let $S \in \mathbb{R}^{H \times W \times 1}$ be the hand-drawn sketch corresponding to the masked object containing visual details of the object to be inpainted. Since the image encoder $E(.)$ is a CNN model, the latent code related to any patch in an image is affected by the pixel values of its neighbors. Hence, directly masking the input image in the pixel space $I_M = M \odot I_{GT}$ is erroneous for obtaining latent representation (where $\odot$ represents element-wise multiplication). A better way to obtain the masked image $I_M$ is to firstly encode $I_{GT}$ to a sequence of latent codebook entries, $z_0$, and then to replace tokens corresponding to masked regions with a special {\tt [MASK]} token. The original image mask $M$ is transformed to obtain the mask for the discrete latent image representation $M_L \in \{0, 1\}^K$ that represents masked tokens in the latent space. It is important to note here that an embedding corresponding to the {\tt [MASK]} token does not exist in the codebook $Z$, but since we aim to represent the image as a sequence of indices, we assign the index $(C+1)$ to this special token. To sum up, we encode the image $I_M$ in discrete latent space as $z_m=(C+1)M_L + (1-M_L) \odot z_0$. 
% \begin{center}
%     \begin{equation}
%         z_m = (C+1)M_L + (1-M_L) \odot z_0.
%     \end{equation} 
% \end{center}
\subsubsection{PDDP: Partial Discrete Diffusion Process} In order to obtain the denoised latent image representation $z_0$ from the noisy representation $z_m$, iterative denoising of $z_m$ is performed until we have a visually plausible $z_0$~\cite{2021VectorQD}. Yet, it does not allow explicit training for the inpainting problems as it does not align with the inference where we have to diffuse from an intermediate state with the desired masked region to obtain $z_0$. In this work, we introduce a novel inpainting model called Partial Discrete Diffusion (PDD), which can be used to train general inpainting models in the discrete latent space. With PDD, we aim to align the forward and backward processes of discrete diffusion for image inpainting. Specifically, we propose a forward process that gradually corrupts $z_0$ to $z_m$ in $T \in \mathbb{N}$ timesteps, i.e., in our formulation, $z_T = z_m$. During inference for inpainting, this means denoising our corrupted latent representation $z_m$ to $z_0$ in $T$ timesteps. The fact that the single-step transition probability for each token at each timestep is independent of each other allows us to incorporate the position of masked regions into the transition distribution $q(z_t \lvert z_{t-1})$ by augmenting it as follows: $M_L \odot v(z_t)^\top\mathbf{Q_t} {v}(z_{t-1}) + (1 - M_L) \odot {v}(z_{t-1})$. Furthermore, the $t$ step transition probability $q(z_t \lvert z_{0})$ can be obtained as follows: $M_L \odot {v}(z_t)^{T}\mathbf{\bar{Q}_t}{v}(z_{0}) + (1 - M_L) \odot {v}(z_{0})$.

\subsubsection{Modelling the Reverse Process} 
\label{subsubsec:reverse_process}
We learn to reverse the partial discrete diffusion process to obtain $z_0$ from $z_m$ iteratively. Typical parameterization of the reverse process comprises of predicting un-normalized log probabilities $\log p_\theta(z_{t-1} \lvert z_t)$. But the recent works \cite{mulinomialdiffusion, 2021DiscreteDiffusion} have found that directly predicting the noiseless target variable $q(z_0)$ results in a better quality of generated images. This formulation is achieved by using the following reparameterization trick, which results from the Markovian nature of the forward discrete diffusion process:
\begin{center}
    \begin{equation}
    \label{eqn:q_rev_through_z0}
        \begin{aligned}
        q(z_{t-1} \lvert z_t, z_{0}) & = \frac{q(z_t \lvert z_{t-1}, z_0) q(z_{t-1}, z_0)}{q(z_t \lvert z_0)}.
        \end{aligned}
    \end{equation}
\end{center}

Building on this, we design our neural network $p_\theta(.)$ to predict the distribution of noiseless target variable $\tilde{z_0}$, estimated at each reverse step conditioned on sketch embeddings $s$. Using this $p_\theta(\tilde{z_0} \lvert z_{t}, s)$, we can compute the one-step reverse transition~\cite{2021VectorQD} using the following equation which combines it with the posterior $q(z_{t-1} \lvert z_t, z_{0})$:
\begin{center} 
    \begin{equation}
    \label{eqn:reparam_eqn}
        p_\theta(z_{t-1} \lvert z_t, z_0, s) = \sum_{\tilde{z_0}}q(z_{t-1} \lvert z_t, z_{0})p_\theta(\tilde{z_0} \lvert z_{t}, s).
    \end{equation}
\end{center} We train the network to minimize the Variational Lower Bound (VLB) objective \cite{sohl15diffusion} for one-step reverse prediction of $q(z_{t-1} \lvert z_{t}, z_0)$ and a denoising objective following \cite{2021DiscreteDiffusion, 2021VectorQD}, which encourages the model to predict a better noiseless $\tilde{z_0}$. The total loss is given by: $\mathcal{L}_0 = \mathcal{L}_{VLB} + \lambda \mathcal{L}_{z_0}$, where $\lambda$ is a hyper-parameter used to balance the contributions from the two losses, VLB loss is defined as:
\begin{equation}
\begin{split}
    \mathcal{L}_{VLB} = & -\log p_\theta (z_0 \lvert z_1, s) \\ 
    & + \sum_{T}D_{KL}( q_\theta (z_{t-1} \lvert z_t, s) \lvert \lvert p_\theta (z_{t-1} \lvert z_t, s))
\end{split}
\end{equation} where $D_KL(x\lvert\lvert y)$ denotes the KL-Divergence between the random variables $x$ and $y$. $\mathcal{L}_{z_0}$ is the denoising objective defined as:
$\mathcal{L}_{z_0} = - \log p_\theta (z_0 \lvert z_t, s)$.

\subsubsection{Conditioning the Reverse Process on Sketches} We use a sketch of the object to be inpainted as a conditioning signal to guide the image inpainting process. The sketch image $S \in \mathbb{R}^{224 \times 224 \times 1}$ represents a rough outline of the input image's missing content, indicating the object's overall shape and pose needs to be inpainted. Recent studies have utilized AdaLN~\cite{layernorm} and AdaGN~\cite{dhariwal2021diffusion} parameters for conditioning the generation process. Yet, considering the intricate nature of the sketches with varying spatial information (i.e., shape and pose), conditioning using AdaLN is inadequate. Thus, we propose incorporating the sketch's visual information into our inpainting model through a simple yet effective method. We first pass the hand-drawn sketch through a sketch encoder $E_s$, which we realize as a ResnNet50~\cite{He2015resnet}. The ResNet50 extracts features from the sketch and produces a feature map $f_s \in \mathbb{R}^{7 \times 7 \times 2048}$. We add 2D-learnable positional embeddings to the feature map $f_s$ to further incorporate the positional information into the sketch features and obtain a final flattened feature map $s \in \mathbb{R}^{49 \times 2048}$ for representing the sketch information for further inpainting by reversing the diffusion process. We linearly project these representations before feeding them to our bi-directional transformer.

\subsubsection{Model Architecture} We propose to realize $p_\theta(.)$ as a bidirectional encoder-only transformer to estimate the distribution $p_\theta(\tilde{z_0} \lvert z_{t}, s)$. As shown in Figure~\ref{fig:model_figure}, our model consists of a sketch encoder $E_s$,  and the diffusion decoder $p_\theta(.)$. We use a ResNet50 \cite{He2015resnet} as a sketch encoder $E_s$ and it takes in a hand-drawn sketch $S \in \mathbb{R}^{W_s \times H_s \times 1}$ and maps it to a set of latent features $s$ after adding positional embeddings. At any timestep $t \in \{1, 2, \ldots, T\}$, the goal of our network $p_\theta$ is to take in $z_t$ and $s$ and predict the distribution $q(z_{t-1} \lvert z_0)$. To condition the denoising step on the sketch embeddings $s$, we concatenate a linear projection of latent representation $s$ with the vector representation of $z_t$, i.e., $v_{z_t}$ and then we feed the concatenated representation $[s;  v_{z_t}]$ through a series of bi-directional transformer blocks to finally predict $p_\theta(\tilde{z_0} \lvert z_t, s)$.
%We provide implementation details in the supplementary material. %The linear projection and concatenation of sketch latents help project the sketch and masked image regions to the same sub-space.

% Each transformer block consists of a bidirectional multi-head self-attention module, a normalization module, and a feed-forward network module, along with appropriate skip connections. To aid the denoising process, the diffusion models are conditioned on the current timestep $t$ to denoise the latent $z_t$ to $z_{t-1}$ \cite{sohl15diffusion}. To incorporate timestep information into the denoising process of $p_\theta$, we augment our model with Adaptive Layer Normalization (AdaLN) to inject the information about the current timestep $t$ into the network. It is implemented as follows: AdaLN$(h, t)$ = $a_t$LN($h$) + $b_t$, where $h$ is the hidden activation and $a_t$ and $b_t$ represent a linear projection of the timestep embedding corresponding to time $t$. This network is optimized using a combination of the variational lower bound loss and denoising loss discussed earlier in this section.

\section{Experiments and Results}
\label{sec:expts}
\begin{figure}[!t]
%https://docs.google.com/drawings/d/1YbIJelJO77NPBoTFdnAt2f6pgErYMFVn7S9fTFvcCUU/edit?usp=sharing
\includegraphics[width=0.48\textwidth]{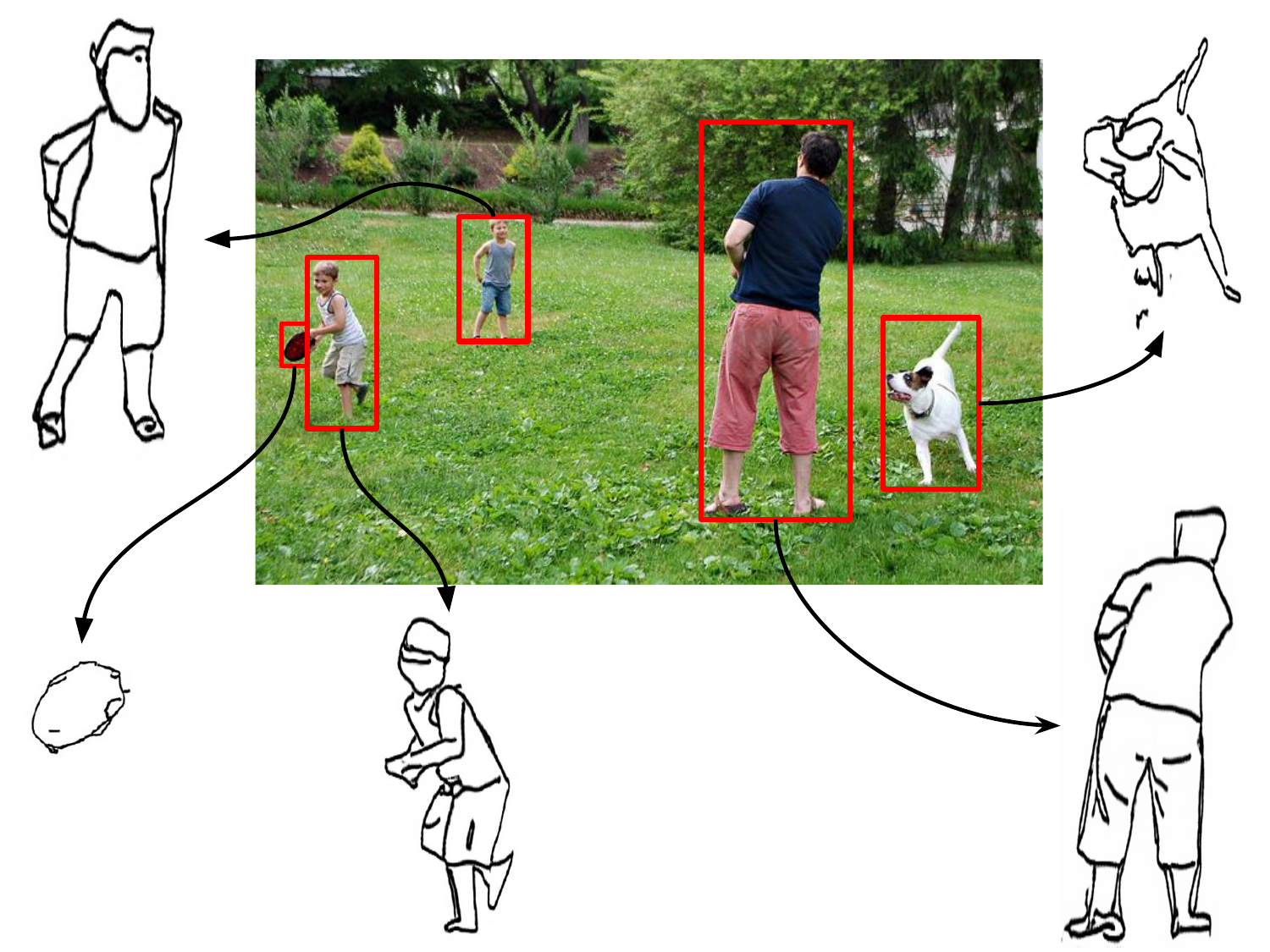}
\caption{\label{fig:data_samples} 
An example of our dataset. We randomly mask a bounding box shown using red color and provide the masked image along with the corresponding sketch as input to the image inpainting method. Please refer to Section~\ref{subsec:dataset} for more details.}
\end{figure}

\subsection{Dataset and Performance Metrics}
\label{subsec:dataset}
We curate a dataset from the widely used MS-COCO dataset~\cite{cocodataset}. We begin by isolating objects of interest and then removing any irrelevant background information by segmenting all objects from the MS-COCO images using available annotations. To improve the resolution of the segmented objects, we employed a pretrained super-resolution method, namely ESR-GAN model~\cite{esrgan}. Finally, we apply PhotoSketching~\cite{photosketching} to produce sketchy versions of the high-quality segmented objects and generate a dataset of images and corresponding object sketches. The resulting dataset contains 860K object sketches in the training set and 36K object-level sketches in the validation set. Please refer to Figure~\ref{fig:data_samples} for a sample image of this dataset. 

Traditional image inpainting metrics like mean squared error, peak signal-to-noise ratio (PSNR), or structural similarity index (SSIM) are not well-suited for sketch-guided image inpainting as the input sketches only provide a rough outline of the missing regions and do not specify the exact content or color palette to be inpainted, resulting in different valid inpainting outputs. Thus, we use the FID score~\cite{2021cleanfid} to measure the quality of the inpainted images and the inpainted region, respectively. We also adopt the LPIPS metric~\cite{2018lpips}, commonly used in recent inpainting works~\cite{2022RePaintIU, 2022GLaMaJS, 2021PaletteID}, to measure the similarity between the inpainted image and ground truth. However, given that these metrics take the entire image into account rather than solely the inpainted region, we additionally present Local-LPIPS and Local-FID score, which measures the LPIPS similarity and FID score between the inpainted region and its corresponding region in the ground truth image. Please note that local metrics are more effective at capturing performance than global metrics when the masked region is small. We evaluate inpainting methods by randomly masking an object using bounding box annotations in 5K images from the MS-COCO validation set, and further conduct user studies to measure the photorealism of the output images and their consistency with the sketch query.

\subsection{Competing Approaches}
To assess our model's performance, we adapted closely related methods by training them on the trainset of our dataset. A brief overview of these approaches is provided below {(i) Sketch-Colorization GAN:} We implement a simple DCGAN~\cite{radford2015dcgan}, which takes a corrupted image $I$ with a sketch $S$ pasted into the missing region through channel-wise concatenation and generates a completed image. The model is then trained to synthesize an image to match the ground-truth image.
\textbf{(ii) DeFLOCNet~\cite{Liu2021DeFLOCNet}} demonstrated state-of-the-art performance on facial attribute editing tasks using partial sketches while generating realistic results. We train the SC-FEGAN baseline with object sketches instead of partial sketches to generate inpainted images. \textbf{(iii) DeepFillv2~\cite{deepfillv2}} is an image inpainting method aimed at filling in missing regions of images with free-form masks. We train this model from scratch by adapting it to our problem setup, where the inpainting is produced by concatenating the sketch with the binary mask and the corrupted image. \textbf{(iv) Palette~\cite{2021PaletteID}} is based on Denoising Diffusion Probabilistic Models. We adapt this model to our proposed setup by training it for the inpainting task by conditioning the inpainting process on the latent representation of the sketch obtained through a ResNet-50 encoder, using the AdaLN layer~\cite{layernorm}.
\textbf{(v) ControlNet~\cite{controlnet23}}  introduces conditional control to the Stable Diffusion ~\cite{2021LatentDiffusion} model. In our early experiments, the pre-trained models demonstrated poor performance for our task because these models are designed for text-to-image generation and not image inpainting. Therefore, we train ControlNet on our dataset by providing a scene image containing masked region, sketch query, and a default caption, ``A photo-realistic image''.

\subsection{Results and Discussion}
\label{sec:results}

    \begin{figure*}[h]
    \begin{center}
    \includegraphics[width=\textwidth]{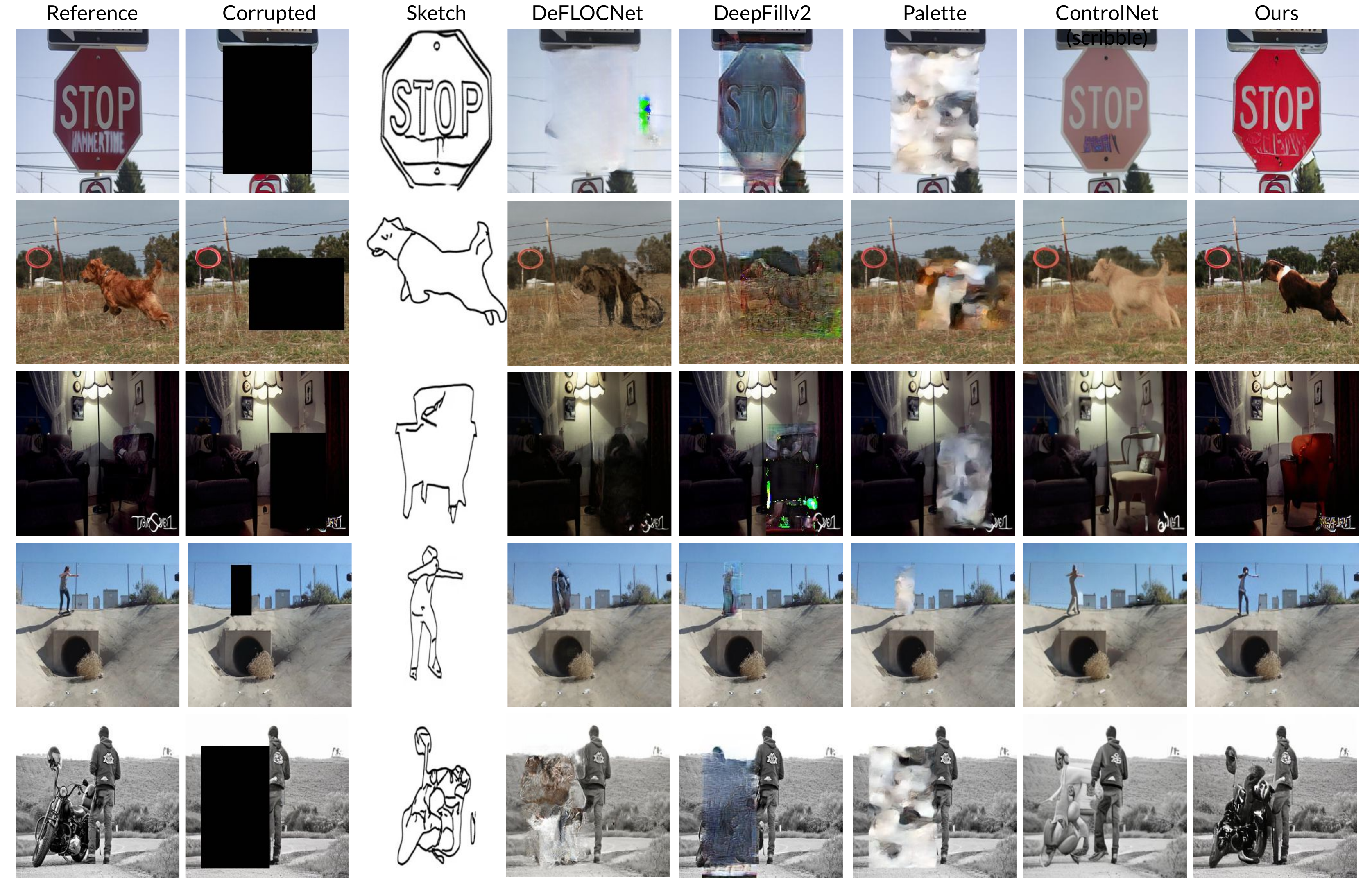}
    \caption{\label{fig:Quali_results} Qualitative comparison of the proposed sketch-guided inpainting method with the competitive baselines. The results show that the proposed model effectively utilizes visual information in the sketch query, producing inpainting results with high visual fidelity and query faithfulness. Refer to Section~\ref{sec:results} for more details.}
\end{center}
    
    \end{figure*}
We conducted experiments to evaluate the models' performance for sketch-guided image inpainting and present the results in Table~\ref{tab:mainResults}. It is evident from the Table that the Sketch-colorization GAN method performs inferior to all other methods. This is because Sketch-colorization GAN is a na\"{\i}ve approach that generates an inpainted image from a corrupted image with a sketch pasted into the missing region. DeFLOCNet and DeepFillv2 also fail to encapsulate the information from crude, object-level sketches. The Palette uses AdaLN~\cite{layernorm} to condition the inpainted region on the sketch, which produces poor results since the conditioning mechanism cannot encapsulate accurate information about the shape and pose of the object from the hand-drawn sketch. The pre-trained ControlNet performs poorly at our proposed task, achieving an FID score of 29.52, which is attributed to the fact that it utilizes a pre-trained text-to-image StableDiffusion model, which makes its output heavily dependent on the text prompt. Therefore, we train a ControlNet from scratch on our curated data, achieving an FID score of 10.77. This is our closest-performing model that produces better visual results than the other approaches. However, in many cases, as seen in Figure~\ref{fig:Quali_results}, the faithfulness of the inpainted region with the provided hand-drawn sketch remains an unresolved issue with this model, too.
In contrast, our proposed technique achieves state-of-the-art performance in sketch-guided image inpainting, as demonstrated by a lower (i.e., better) FID score in Table~\ref{tab:mainResults}. It is worth noting that the Palette, ControlNet, and our framework are based on diffusion models. Palette performs diffusion in latent space, and inpainting is conditioned only through a single sketch embedding.
On the other hand, ControlNet uses diffusion in the continuous space and StableDiffusion's U-Net encoder~\cite{ronneberger2015u} to guide the pre-trained StableDiffusion model in generating the inpainted images. Our proposed solution exploits discrete diffusion and benefits from a more robust sketch-conditioning mechanism where the features are first extracted from a ResNet50 model and then combined with image features in the self-attention blocks of the transformer. %FID for our proposed method v.s. \textbf{10.77} for the ControlNet model). The proposed method uses the full sketch feature map and a powerful multimodal transformer architecture to generate an inpainted image with high visual fidelity, semantic consistency, and sketch faithfulness, which is reflected by the model's performance when measured using the Local-LPIPS metric.
\begin{table}[!t]
    \footnotesize
    \centering
    {
      \begin{tabular}{l r r r r}
        \toprule
        \textbf{Method} &  \textbf{FID $(\downarrow)$} &  \textbf{LPIPS $(\downarrow)$} & \textbf{LLPIPS $(\downarrow)$} & \textbf{LFID $(\downarrow)$}   \\
        \midrule
        Sketch-coloring GAN & 37.23 & 0.79 & 0.98 & 152.64 \\
        DeFLOCNet~\cite{Liu2021DeFLOCNet} & 30.68 & 0.17 & 0.57 & 76.16\\
        DeepFillv2~\cite{deepfillv2} & 27.19 & 0.16 & 0.55 & 105.49 \\
        Palette~\cite{2021PaletteID} & 25.87 & 0.14 & 0.53 & 98.56 \\
        % ControlNet (zero-shot)~\cite{controlnet23} & 29.52 & 0.19 & 0.56\\
        ControlNet~\cite{controlnet23} & \underline{10.77} & \underline{0.11} & \underline{0.49} & \underline{21.98}\\
        \midrule
        Ours & \textbf{7.72} & \textbf{0.11} & \textbf{0.42} & \textbf{21.91} \\
       \bottomrule
      \end{tabular}
      }

      \caption{Performance comparison of the proposed model on the curated MS-COCO dataset. Lower is better (see Section~\ref{sec:results}). LLPIPS and LFID denote ``Local LPIPS'' and ``Local FID'', respectively. Bold and underlined numbers refer to the best and second-best performances for their respective metrics.}
      \label{tab:mainResults}
\end{table}

The qualitative results of our model in comparison to competitive approaches for the proposed task are shown in Figure~\ref{fig:Quali_results}. Our proposed model outperforms all the competent approaches and successfully utilizes the visual shape and pose information from the query sketch to generate high visual fidelity, semantic consistency, and faithfulness to the provided sketch when inpainting the missing region. We omit the results of sketch-colorization GAN because of poor quantitative performance.

\begin{table}[t!]
\centering
\small
\begin{tabular}{lccccc}
\toprule
Timesteps (T) & 1~~~~~     & 2~~~~~     & 10~~~~~    & 25~~~~~    & 50~~~~~    \\
\midrule
FID           & 12.55~~~~~ & 10.40~~~~~ & 8.17~~~~~  & 7.89~~~~~  & 7.72~~~~~  \\
LPIPS         & 0.124~~~~~ & 0.110~~~~~ & 0.109~~~~~ & 0.109~~~~~ & 0.107~~~~~ \\
LLPIPS   &   0.498~~~~~    &  0.480~~~~~      &   0.438~~~~~    &  0.429~~~~~   & 0.414~~~~~\\
\bottomrule
\end{tabular}
\caption{The analysis of the effect of the number of inference steps on the quality of the inpainted images (refer to Section~\ref{subsec:ablation}).}
      \label{tab:diffStepsAblation}
\end{table}

\begin{table}[t!]
% \scriptsize
\centering
\begin{tabular}{lr}
\toprule
Method & User Preference (\%)\\
\midrule
DeFLOCNet~\cite{Liu2021DeFLOCNet} & 2.72\\
DeepFillv2~\cite{deepfillv2} & 2.72\\
Palette~\cite{2021PaletteID} & 0.01\\
ControlNet~\cite{controlnet23} & 25.45\\\midrule
Ours & \textbf{68.54}\\
\bottomrule
\end{tabular}
\caption{\label{tab:hum_eval} The study involved 50 masked images randomly selected from the validation split of our dataset. A group of 22 human participants were presented with the inpainted images generated by our method and competing approaches. They were then asked to express their preferences, focusing on photorealism.}
\end{table}
%\vspace{-5pt}
% \begin{table}[!h]
%     \centering
%     {
%       \begin{tabular}{l c c c}
%         \toprule
%         \textbf{Inference Steps} &  \textbf{FID} &  \textbf{LPIPS} & \textbf{Local LPIPS}  \\
%         \midrule
%         1 & 12.55 & 0.124 & - \\
%         2 & 10.40 & 0.110 & - \\
%         10 & 8.17 & 0.109 & - \\
%         25 & 7.89 & 0.109 & - \\
%         50 & 7.72 & 0.107 & 0.451\\
%        \bottomrule
%       \end{tabular}
%       }

%       \caption{The analysis of the effect of the number of inference steps on the quality of the inpainted images (Refer to Section~\ref{subsec:ablation}).}
%       \label{tab:diffStepsAblation}
% \end{table}
% \vspace{-15pt}
\noindent\textbf{Ablation study:}
\label{subsec:ablation}
Recent studies~\cite{2021DiscreteDiffusion, 2021LatentDiffusion, 2021VectorQD} have shown that the quality of images generated by diffusion models is affected by the number of diffusion timesteps ($T \in \mathbb{N}$). To study this, we conducted a study to measure the effect of the number of inference timesteps on the quality of inpainted images. The results in Table~\ref{tab:diffStepsAblation} demonstrate that the quality of the generated inpainted images increases as the number of inference steps increases with the highest FID of \textbf{7.72} for 50 steps. Even the images generated in two timesteps have a better quality than the closest baseline model (FID score of \textbf{10.77} for ControlNet v.s. \textbf{10.40} FID for our method).\\
\noindent\textbf{User Study:} In addition to quantitative evaluations, we also conducted a subjective assessment of the proposed inpainting method using a human preference metric. To carry out the assessment, we randomly selected 50 masked images and inpainted them using four top-performing competitive approaches and our proposed framework. We recruited 22 human users to participate in the evaluation. Each user was presented with all sets of masked images and corresponding inpainted versions. The subjects were then instructed to carefully examine each set and choose the inpainted image they perceived as the most natural-looking. The human preference evaluation results, indicating the participants' choices, are reported in Table~\ref{tab:hum_eval}. As shown, 68.54\% of the time, the users preferred the inpainted results generated by our method. This analysis indicates the naturalness and visual fidelity of the generation. Additionally, we performed another user study to quantitatively evaluate how well the inpainted region of our and the baseline models align with the provided user sketch. We randomly select 20 images from the validation split and show 22 human subjects the inpainting results produced by competing baselines and our method along with the query sketch, and ask them to rate the consistency and alignment of the inpainted region with the given sketch from 1 (poor alignment and consistency) and 5 (best alignment and consistency). The results in Table~\ref{tab:hum_eval2} indicate the superior performance of our proposed method.
% \vspace{-18pt}
% The reparameterization formulation of $p_\theta(z_{t-1} \lvert z_t, s)$ described in Equation \ref{eqn:reparam_eqn} allows us to skip some steps during inference to speed up the process. Specifically, we can skip one timestep after each step to perform inference in 25 steps and skip four steps to perform inference in 10 steps. We present the results of our ablation study on inference steps in Table \ref{tab:diffStepsAblation}.

% Recent works \cite{sohl15diffusion, ho2020ddpm, dhariwal2021ddim, mulinomialdiffusion, 2021DiscreteDiffusion, 2021LatentDiffusion, 2021VectorQD} have found out that the quality of samples generated using diffusion models is dependent on the number of diffusion timesteps $T \in \mathbb{N}$. Therefore, we ablate the choice of total timesteps at the time of inference to measure how the quality of inpainted images changes with the number of inference timesteps. The formulation of $p_\theta(z_{t-1} \lvert z_t, s)$ as described by Equation \ref{eqn:reparam_eqn} allows us to perform inference by skipping some steps to achieve a faster inference. Specifically, we can skip one timestep at each after each step to perform inference in 25 steps and skip four steps to perform inference in 10 steps. We report the performance on ablated inference steps in Table \ref{tab:diffStepsAblation}.

% The quality of sketches used to guide the inpainting process plays a significant role in determining the overall semantics and texture of the object to be inpainted in the corrupted region.
\noindent\textbf{Limitations:} Our method achieves state-of-the-art performance in sketch-guided image inpainting. However, there is significant room for improving the visual quality of the inpainted images. Our work represents a small step towards object-level sketch-guided image inpainting. One area to enhance is our sketch information embedding, which currently uses a straightforward ResNet50 encoder for extracting embeddings from rasterized sketches. Future research could explore more sophisticated sketch embeddings capturing stroke-level details. Another area of exploration involves refining the conditioning mechanisms that merge sketch embeddings with image representations to synthesize the inpainted image. Furthermore, we aim to develop diffusion models for their generative capabilities and leverage transformer models for their robust modeling capabilities, building upon the discrete diffusion process. %Recent advances like DiT~\cite{peebles2023scalable} illustrate the potential of transformers within a continuous diffusion framework for application in this domain.

\begin{table}[!t]
% \scriptsize
\centering
\begin{tabular}{lr}
\toprule
Method & Consistency Score (mean$\pm$std)\\
\midrule
DeFLOCNet~\cite{Liu2021DeFLOCNet} & 1.40 $\pm$ 0.65\\
DeepFillv2~\cite{deepfillv2} & 2.24 $\pm$ 1.07 \\
Palette~\cite{2021PaletteID} & 1.09 $\pm$ 0.33 \\
ControlNet~\cite{controlnet23} & 3.75 $\pm$ 1.20 \\
\midrule
Ours & \textbf{4.34 $\pm$ 0.77}\\
\bottomrule
\end{tabular}
\caption{\label{tab:hum_eval2} The study involved 20 masked images and corresponding sketch queries randomly selected from the validation split of our dataset. A group of 22 human participants were presented with the inpainted images generated by our method and competing approaches. They were to score the consistency of the inpainted region with the sketch on a scale of \textbf{1 (poor) to 5 (best)}.}
\end{table}

%\section{Limitations and Future Work}

%However, collecting stroke-level sketch data for objects in the image can be laborious and resource-intensive.
\section{Conclusion}
In this work, we investigated sketch-guided image inpainting, where a query sketch and non-missing regions of the image provide cues for filling in the missing regions. The proposed approach alleviates the problem of limited control over inpainted objects in traditional image inpainting and text-to-image inpainting, thereby making it more practical for image manipulation applications.  Despite challenges such as the significant domain gap between hand-drawn sketches and images, our proposed approach achieved state-of-the-art results and generated photo-realistic objects that fit the context in terms of the shape and pose of the object in the provided sketch. Both quantitative and qualitative analyses demonstrate that our approach significantly outperforms other relevant approaches.

{
    \small
    \bibliographystyle{ieeenat_fullname}
    \bibliography{main}
}

\maketitlesupplementary
\setcounter{section}{0}
\setcounter{figure}{0}
\setcounter{table}{0}
\section{Qualitative Case Studies}
\label{sec:qual_ablations}
\subsection{\textbf{How does the proposed approach work for sketch queries of varying levels of styles and details?}}
\begin{figure}[!t]
\centering
\includegraphics[width=0.45\textwidth]{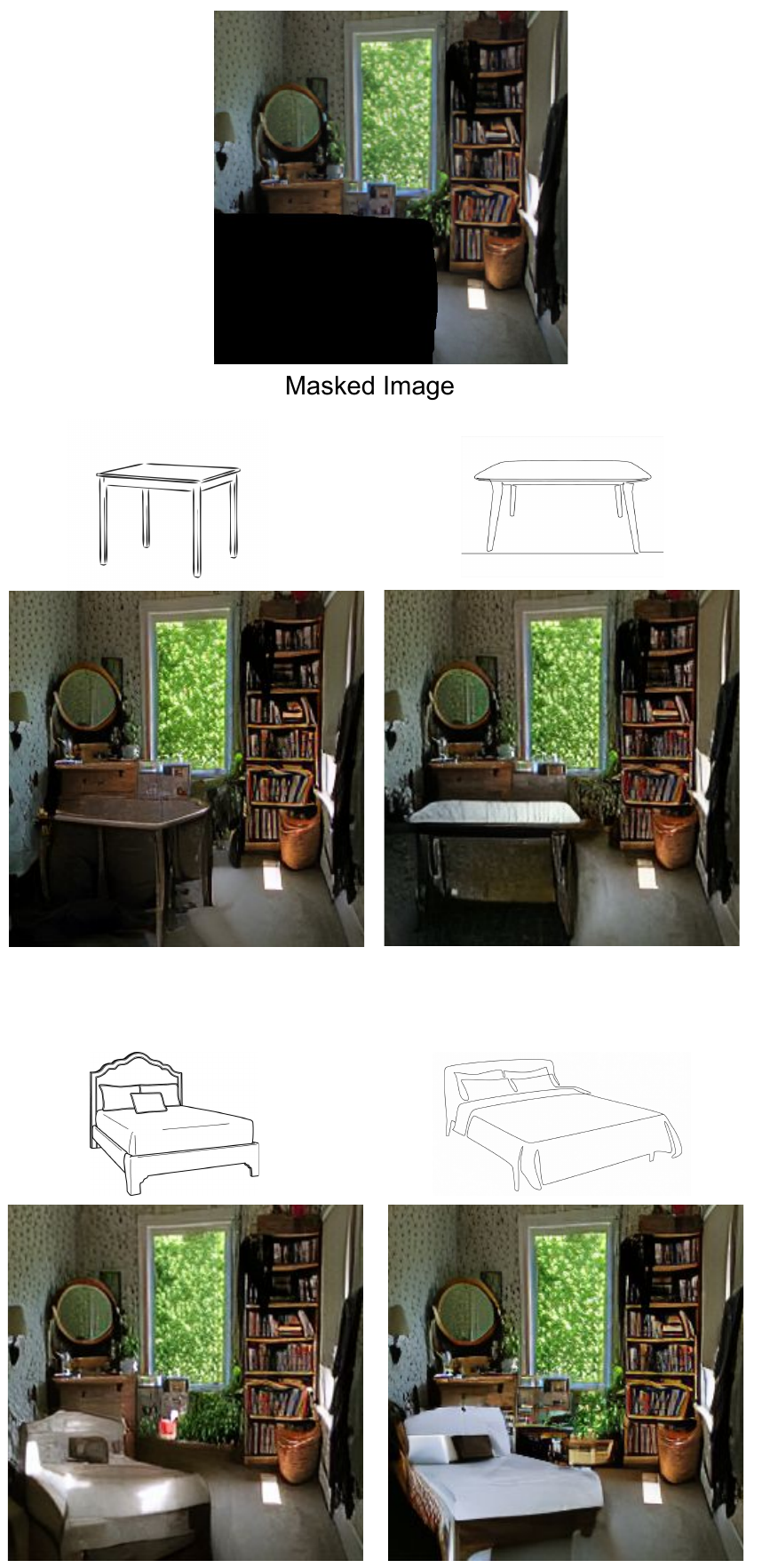}
\caption{\label{fig:hand_drawn_abl} Evaluation of our model with non-synthetic sketches with varying styles and levels of detail. Sketches are sourced from the web.}
\end{figure}

In Figure~\ref{fig:hand_drawn_abl}, we conduct a qualitative study to evaluate the effectiveness of our proposed approach in handling out-of-domain non-synthetic sketches with diverse levels of detail and style. Specifically, we mask a portion of the bedroom image and perform inpainting using two sketches depicting a table and two sketches representing a bed. These sketches are selected such that they exhibit significantly different styles and levels of detail. The results demonstrate the robustness of our approach to varying levels of styles.

% we qualitatively study the strength of our proposed approach in handling out-of-domain non-synthetic sketches of varying levels of detail and style. Specifically, we mask a portion of the bedroom image and inpaint it using two sketches of a table and two sketches of a bed. The results demonstrate the robustness of our approach to varying levels of styles.

\subsection{\textbf{How does the proposed approach work for partial sketch queries?}}
Our research is centered on \textbf{object-level sketch-guided} image inpainting. Unlike methods that work on a sketch or partial stroke guidance in sketch manipulation tasks, our work is designed for whole-object sketches, thus impacting its performance on partial sketch query due to our model's training paradigm. During training, our model is expected to reconstruct the masked region using the sketch of the whole object. So, the model interprets that every input sketch represents the sketch of a full object, which needs to be inpainted fully into the masked region. So, at the test time, when we mask regions and provide a partial sketch, the model interprets that the provided sketch represents the whole object and does not provide appropriate inpainting. This may be addressed by training the proposed model using the partial sketch. However, this substantially digresses from the original goal of this work and hence is beyond the scope of this paper.

\subsection{Inpainting using Edgemaps}
In some cases, the sketches generated by the proposed method exhibit abstract representations with limited object details. To address this, we explored an alternative approach by utilizing the edge map of the masked image obtained by applying a Laplacian kernel. The quantitative results of inpainting using the edge-map as a query are presented in Table~\ref{tab:edgemap_comp}. The generated edge maps often capture more intricate object details, contributing to improved inpainting results when compared to inpainting using sketch queries. We continued training the sketch-guided model from the 150th epoch by conditioning it on edge maps until convergence.

\begin{table}[!t]
\centering
\small
\begin{tabular}{lcc}
\toprule
Guidance & Synthetic Sketch     & Edge Map \\
\midrule
FID           & 7.72 & 7.59  \\
LPIPS         & 0.107 & 0.106 \\
\bottomrule
\end{tabular}
\caption{\label{tab:edgemap_comp} Comparison of Inpainting Results using Edge-Maps as a query/guidance.}
\end{table}

\begin{figure*}[!t]
\centering
\includegraphics[width=0.95\textwidth]{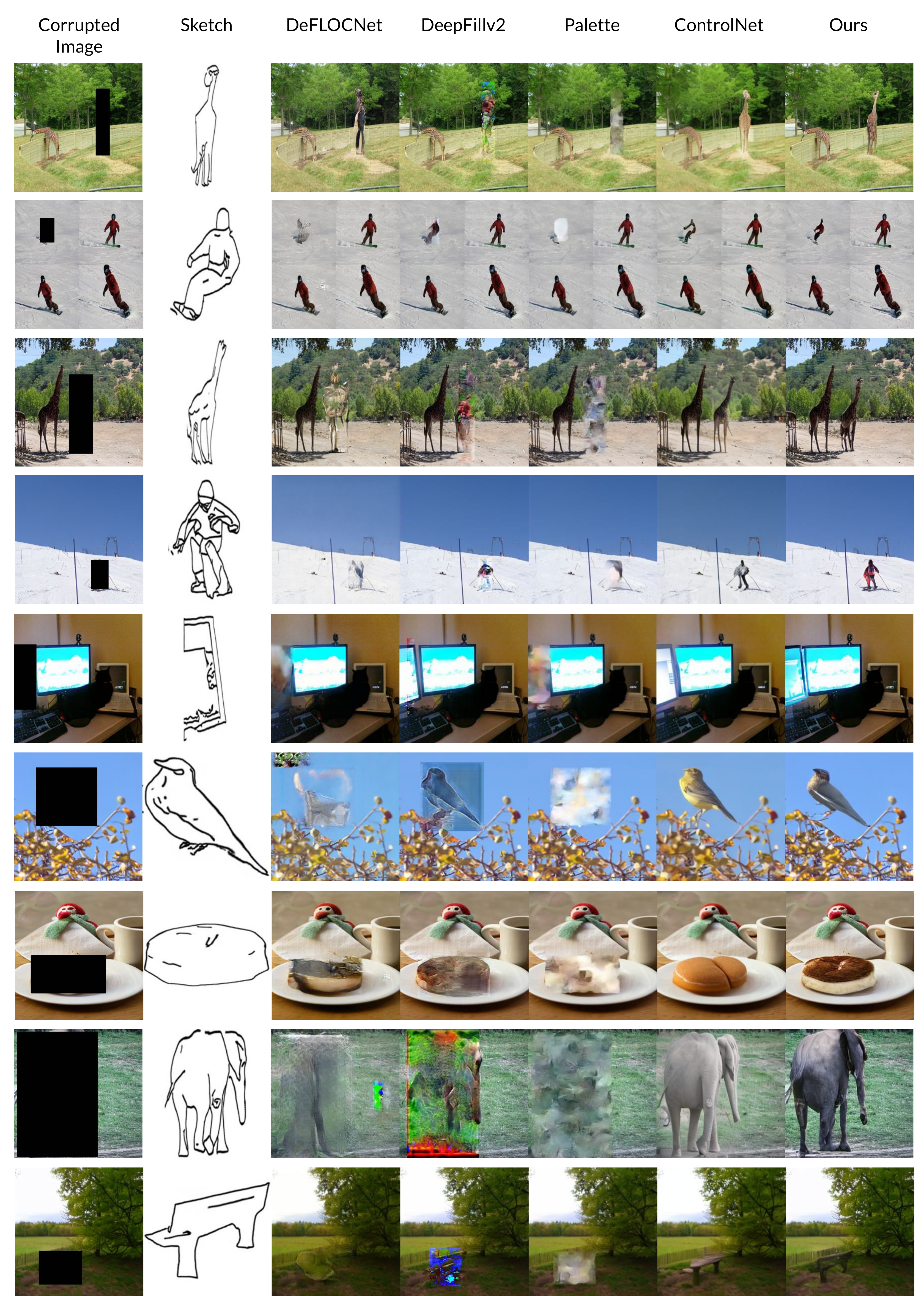}
\caption{\label{fig:supp_qual_result} Additional results for the qualitative comparison of the proposed method with competing baselines.}
\end{figure*}

\noindent\textbf{Additional Qualitative Results:}
In Figure~\ref{fig:supp_qual_result}, we show more qualitative results to contrast the results of our baselines with our method. %We observe that our model consistently produces better-inpainted images when compared to the competing methods.

%https://docs.google.com/drawings/d/13zn5upzyk7pA_ARFn3-K6N0G8DhCa-G8yKQE06DeBHc/edit

\end{document}